\pdfoutput=1

\documentclass[11pt]{article}

\usepackage[]{acl}

\usepackage{times}
\usepackage{latexsym}

\usepackage[T1]{fontenc}

\usepackage[utf8]{inputenc}

\usepackage{microtype}

\usepackage{inconsolata}

\usepackage{graphicx}

\usepackage{bm}
\usepackage{amsmath}
\usepackage{url}
\usepackage{graphicx}
\usepackage{float} 
\usepackage{subfigure}
\usepackage{multirow}
\usepackage{makecell}
\usepackage{xpinyin}
\usepackage{enumerate}

\usepackage{xcolor}
\usepackage{bbding}
\usepackage{soul}
\newcolumntype{L}[1]{>{\raggedright\let\newline\\\arraybackslash\hspace{0pt}}m{#1}}
\setul{2pt}{2pt}
\usepackage{booktabs}
\usepackage{amsmath,amsfonts}
\usepackage[utf8]{inputenc}
\usepackage{cleveref}

\crefname{section}{§}{§§}
\Crefname{section}{§}{§§}
\usepackage{enumitem}
\setenumerate[1]{itemsep=0pt,partopsep=0pt,parsep=\parskip,topsep=5pt}
\setitemize[1]{itemsep=0pt,partopsep=0pt,parsep=\parskip,topsep=5pt}
\setdescription{itemsep=0pt,partopsep=0pt,parsep=\parskip,topsep=5pt}
\usepackage{arydshln}

\usepackage[linesnumbered,ruled,vlined]{algorithm2e}

\title{Beyond Token-Level Routing: Task-Guided and Context-Aware Routing in Sparse MoE for Neural Machine Translation}
\title{THOR-MoE: Hierarchical Task-Guided and Context-Responsive Routing for Neural Machine Translation}

\author{Yunlong Liang, \ Fandong Meng\thanks{ \ \ Corresponding author.},  \ Jie Zhou \\
Pattern Recognition Center, WeChat AI, Tencent Inc \\ 
\texttt{\{yunlonliang,fandongmeng,withtomzhou\}@tencent.com} \\
}

\begin{document}
\maketitle
\begin{abstract}
The sparse Mixture-of-Experts (MoE) has achieved significant progress for neural machine translation (NMT). However, there exist two limitations in current MoE solutions which may lead to sub-optimal performance: 1) they directly use the task knowledge of NMT into MoE (\emph{e.g.}, domain/linguistics-specific knowledge), which are generally unavailable at practical application and neglect the naturally grouped domain/linguistic properties; 2) the expert selection only depends on the localized token representation without considering the context, which fully grasps the state of each token in a global view. 
To address the above limitations, we propose THOR-MoE via arming the MoE with hierarchical task-guided and context-responsive routing policies. Specifically, it 1) firstly predicts the domain/language label and then extracts mixed domain/language representation to allocate task-level experts in a hierarchical manner; 2) injects the context information to enhance the token routing from the pre-selected task-level experts set, which can help each token to be accurately routed to more specialized and suitable experts. Extensive experiments on multi-domain translation and multilingual translation benchmarks with different architectures consistently demonstrate the superior performance of THOR-MoE. Additionally, the THOR-MoE operates as a plug-and-play module compatible with existing Top-$k$~\cite{shazeer2017} and Top-$p$~\cite{huang-etal-2024-harder} routing schemes, ensuring broad applicability across diverse MoE architectures. For instance, compared with vanilla Top-$p$~\cite{huang-etal-2024-harder} routing, the context-aware manner can achieve an average improvement of 0.75 BLEU with less than 22\% activated parameters on multi-domain translation tasks.
\end{abstract}
\section{Introduction}
The rapid advancement of neural machine translation (NMT; ~\citet{zhang-etal-2020-improving,DBLP:journals/corr/abs-2010-11125}) has been significantly propelled by sparse Mixture-of-Experts (MoE) architectures~\cite{shazeer2017,lepikhin2021gshard,fedus2022switch}, which increase model capacity while maintaining computational efficiency through conditional computation. In the literature, existing work can be roughly classified into two categories: (1) some studies mainly focus on how to effectively incorporate task information (\emph{e.g.,} domain/linguistics-specific knowledge) into MoE models designing task-specific modules~\cite{pham-etal-2023-task,jiang-etal-2024-med}; (2) another line of work concentrates on how to improve the training and inference efficiency via reducing the activated experts in MoE~\cite{jawahar-etal-2023-automoe,elbayad-etal-2023-fixing,wang2024hmoeheterogeneousmixtureexperts,zeng-etal-2024-adamoe}.
Specifically, in (1), \citet{kudugunta-etal-2021-beyond-distillation} routes input examples to different experts based on translation language representations. \citet{li-etal-2023-mmnmt} explores multiple language-group-specific routers to incorporate language group knowledge into models. \citet{zhao2024sparse} takes one step further by designing hierarchical language-guided token routing. Although intuitive and effective, they heavily rely on explicit task-specific knowledge (\emph{e.g.}, domain or language labels), which is often unavailable in real-world scenarios. This reliance not only restricts generalization but also overlooks the intrinsic hierarchical grouping of domain and linguistic properties inherent in multilingual or multi-domain translation tasks. In (2), \citet{zhao-etal-2024-hypermoe} distills the knowledge of unselected experts to the selected one and thus reduces the number of experts without impacting performance; \citet{li-etal-2023-adaptive-gating} and \citet{huang-etal-2024-harder} decrease the number of experts by designing a threshold based on expert probability distribution; \citet{zhao2024sparse} adapts the number of experts by the difficulty of language. Generally, the global context grasp the overall situation and might know whether each token is difficult or not~\cite{gloeckle2024betterfasterlarge}. However, previous routing strategies select experts solely based on localized token-level representations, neglecting the broader contextual dependencies that govern token interactions. This myopic view limits the model’s ability to allocate experts optimally in a globally coherent manner. 

In this work, to address the above issues, we propose a innovative hierarchical context-responsive routing (THOR-MoE) framework for neural machine translation. Specifically, the THOR-MoE first predicts which domain/language the input belongs to and then extracts the mixed task\footnote{Note that the task generally denotes the domain/language-specific knowledge in this work.} representation to allocates experts in a hierarchical manner. It is not only flexible in practice but also effectively infuses the domain knowledge into the task-level routing via encouraging experts to specialize in certain domains/languages. Furthermore, before each token routing, the THOR-MoE injects the context information to help accurately assign the experts from the pre-selected task-level experts set for each token in a global perspective. This enables the model to adaptively capture evolving contextual dependencies, such as long-range syntactic structures or discourse coherence. Crucially, the THOR-MoE operates as a plug-and-play module compatible with existing Top-$k$~\cite{shazeer2017} and Top-$p$~\cite{huang-etal-2024-harder} routing schemes, ensuring broad applicability across diverse MoE architectures.  

We validate our proposed THOR-MoE framework on the commonly-used multi-domain NMT benchmark~\cite{aharoni-goldberg-2020-unsupervised} and multilingual NMT benchmark~\cite{zhang-etal-2020-improving}, which contains 5 domains and 16 languages, respectively. Extensive experiments show that the hierarchical context-responsive routing is pivotal for unlocking the full potential of MoE systems. 

In summary, our main contributions are:

\begin{itemize}[leftmargin=*]

\item We propose two components to expand MoE: (i) a hierarchical design to enable language- and domain-specific expert routing; (ii) enabling the use of context information during expert selection, which is important for translation.

\item Extensive experiments on both multi-domain and multilingual translation consistently show the effectiveness and generalization of THOR-MoE. For example, our model achieves consistent improvements of an average improvement of 0.75 BLEU with less than 22\% activated parameters over vanilla Top-p routing in multi-domain translation tasks.

\end{itemize}

\section{Background}

\subsection{Neural Machine Translation}

Given an input sentence in the source language $X$$=$$\{x_i\}_{i=1}^{|X|}$, the goal of the model is to produce its translation in the target language $Y$$=$$\{y_i\}_{i=1}^{|Y|}$. The conditional distribution of the model is:
\begin{equation}
\label{eq:nmt}
    \mathcal{L}_{\text{NMT}} = -\sum_{t=1}^{|Y|}\mathrm{log}(p(y_t|X, y_{1:t-1})),
\end{equation}
where $y_{1:t-1}$ is the partial translation. 

\subsection{Sparse Mixture-of-Experts}

The vanilla MoE model can be seen the variant of Transformer model via replacing the feed-forward (FFN) sub-block of the transformer block with MoE layer, in which per token selects fixed number of experts~\cite{DBLP:conf/iclr/LepikhinLXCFHKS21}. In each MoE layer, there are $N$ experts, denoted as $\mathcal{E} = \{e_1, e_2, \ldots, e_N\}$. An input $\mathbf{x}$ is dispatched to these experts, and the output of the MoE layer is computed as the weighted average of the outputs from the experts:

\begin{equation}
    \textrm{MoE}(\mathbf{x}) = \sum_{i=1}^N g_i(\mathbf{x}) * e_i(\mathbf{x}),
\end{equation}
where $g_*(\mathbf{x})$ is computed by a router that predicts the contribution of each expert to the final output. Given the computing efficiency, the MoE assigns each token to limited experts (\emph{e.g.}, 1 or 2). 

To derive $g_*(\mathbf{x})$, we generally compute the probability $\mathbf{P}$ of each expert being selected for the input $\mathbf{x}$ as follows:
\begin{equation}
    \mathbf{P} = \textrm{Softmax}(\mathbf{W_r} \cdot \mathbf{x}^T),
    \label{eq:P}
\end{equation}
where $\mathbf{W_r} \in {N \times d}$ serves as a learnable parameter and $d$ denotes the dimension of the input $\mathbf{x}$. The vector $\mathbf{P}$, with a size of $N$, encapsulates the probabilities associated with the selection of each expert. And $P_i$ indicates the likelihood of choosing the $i^{th}$ expert $e_i$ to calculate the input $\mathbf{x}$. 
\textbf{\begin{figure*}[t]
    \centering
    \includegraphics[width=0.99\textwidth]{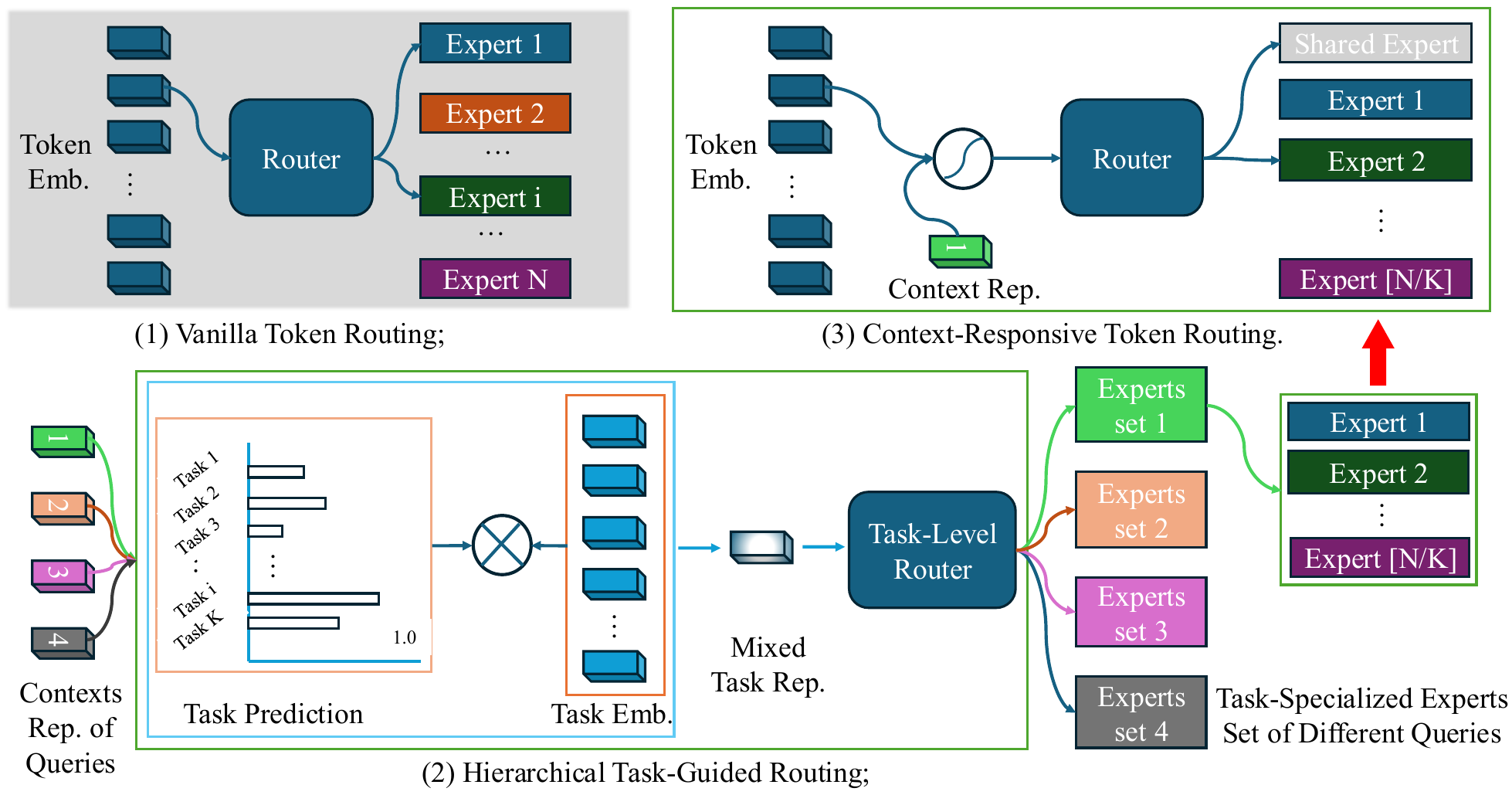}
    \caption{The overview of the proposed THOR-MoE. 1) vanilla token routing; 2) hierarchical task-guided routing; 3) context-responsive token routing. The `Emb.' and `Rep.' denotes the embedding and representation, respectively. The hierarchical manner denotes that the task-guided routing firstly selects different task-specific experts sets for different queries (\emph{e.g.}, $\mathcal{E}_t$ for query 1). Then the context-responsive token routing assigns experts from $\mathcal{S}^t$ for each token in query 1. 
    }
    \label{fig.2}
\end{figure*}}

\subsubsection{Top-$k$ Routing}
\label{topk}
Top-$k$ routing~\cite{DBLP:conf/iclr/ShazeerMMDLHD17,DBLP:conf/iclr/LepikhinLXCFHKS21, DBLP:conf/iclr/Zuo00KHZGZ22} entails the selection of the top $k$ experts, which correspond to the $k$ highest probabilities within the vector $\mathbf{P}$. Then, the probabilities of the chosen experts are re-scaled through normalization, while the probabilities of the non-selected experts are set to zero, signifying their deactivation. The resultant computation for $g_*(\mathbf{x})$ proceeds as below:
\begin{equation}
	g_i(\mathbf{x}) = 
        \begin{cases}
	\frac{P_i}{\sum_{j \in \textrm{TopK}(\mathbf{P})} P_j}, &\ i \in \textrm{TopK}(\mathbf{P})\\
	0, &\ i \notin \textrm{TopK}(\mathbf{P})
        \end{cases}%
\end{equation}
where $\textrm{TopK}(\mathbf{P})$ returns the indices of the k largest elements in $\mathbf{P}$. 


\subsubsection{Top-$p$ Routing}
\label{topp}
Top-$p$ routing is initially proposed by~\cite{li-etal-2023-adaptive-gating,huang-etal-2024-harder}, which aims to overcome the drawback of Top-$k$ activating the fixed number of experts overlooking the variability in difficulty across different inputs. We refer to readers to~\citet{huang-etal-2024-harder} for more details. 

Formally, Top-$p$ first sorts the elements in $\mathbf{P}$ from highest to lowest, returning in a sorted index list $I$. Subsequently, we identify the minimal subset of experts for which the aggregate probability surpasses the threshold $p$. The count of selected experts, denoted as $t$, is then determined by:
\begin{equation}
    t = \mathop{\textrm{arg min}}_{k \in \{1..., N\}} \ \ \sum_{j <= k}{P_{i,j}}  \geq p,
    \label{eq:t}
\end{equation}
where $p$ is the threshold that controls how confident the model should be when stopping adding more experts. $p$ is a hyper-parameter whose range is from 0 to 1. The higher the $p$ is, the more experts will be activated.

In dynamic routing mechanism, the calculation of $g_*(\mathbf{x})$ is:
\begin{equation}
	g_i(\mathbf{x}) = 
        \begin{cases}
	P_i &\ e_i \in S \\
	0, &\ e_i \notin S
        \end{cases}%
\end{equation}
where $S$ is the set of selected experts controlled by $t$ in Equation~\ref{eq:t}:
\begin{equation}
    S = \{e_{I_1}, e_{I_2} ... e_{I_t}\}.
    \label{eq:S}
\end{equation}

\section{THOR-MoE}
In this section, we introduce the proposed hierarchical context-responsive routing framework as shown in Figure~\ref{fig.2}, including two components: \emph{hierarchical task-guided routing} (\autoref{sec:dgr}) and \emph{context-responsive token routing} (\autoref{sec:ces}). Finally, we describe the \emph{training objectives} (\autoref{sec:tt}) in detail.

\subsection{Hierarchical Task-Guided Routing}
\label{sec:dgr}
Our hierarchical task-guided routing consists of three stages, as shown in 2) of~\ref{fig.2}. The first stage is task prediction, which aims to automatically obtain the task knowledge by predicting which domain/language the input belongs to. The second stage is to inject the mixed task representation to allocates experts in a hierarchical task-guided manner (third stage). 

\noindent\textbf{Task Predictor.} During real testing for a query, it is hard to reach its domain or language category. Therefore, to automatically obtain the domain knowledge or language knowledge, we add a special token [CLS] in front of the input. Then, it represents the global sentence information and thus can effectively reflect which domain/language the input belongs to~\cite{devlin-etal-2019-bert}. Finally, we transform the representation $\mathbf{H}^{cls}$ into a fixed-size vector through $\rm{Maxpooling}$ and apply a fully-connected layer to predict suitable labels:
\begin{equation}
  \label{domain_softmax}
  \begin{split}
  \mathbf{H}^{max} &= {\rm Maxpooling}(\mathbf{H}^{cls}),\ \mathbf{H}^{max} \in \mathbb{R}^{d}, \\
    \mathcal{P}^{t} &= {\rm Softmax}(\mathbf{W}^p\mathbf{H}^{max}), 
  \end{split}
\end{equation}
where $\mathbf{W}^p \in \mathbb{R}^{K \times d}$ is trainable weight and $K$ denotes the number of domains or languages and $d$ is the model dimension.

\noindent\textbf{Mixed Task Representation.} Theoretically, we can directly take the predicted task label to extract corresponding task representation. However, in some cases in practice, it is hard to judge since some sentences belongs to multiple domains. In language cases, there are some code-mixed sentences, which is also confused to decide. Therefore, directly use the given label in existing work or the predicted label cannot accurately help model select the specialized experts for such cases and thus limits their potential, as shown in~\autoref{ITP}. To address this issue, we use the mixed task representation as follows: 
\begin{equation}
  \label{gate}
  \mathbf{E}_p = \sum(\mathcal{P}^t \cdot \mathbf{EMB1}), \ \mathbf{E}_p \in \mathbb{R}^{d},
\end{equation}
where $\mathbf{E}_p$ denotes the mixed task representation generated by the weighted sum of the task distribution $\mathcal{P}$ (predicted by the task predictor as Eq.~\ref{domain_softmax}) and the task parameter matrix $\mathbf{EMB1}$. The dimension of $\mathbf{EMB1}$ is $num$ (number of tasks/language groups) * $d$ (embedding dimension), which is randomly initialized and then optimized with the training objective. 

\noindent\textbf{Hierarchical Task-Guided Routing.} With the mixed task representation, we design a task router ${g}^t$ at the task level. In each MoE layer, ${g}^t$ takes the mixed task representation $\mathbf{E}_p$ as input and outputs a task-dependent expert vector. Then, based on it, we use $\textrm{TopK}$ function to select task-specific candidate experts as $\mathcal{S}^t$ from all experts. Finally, based on the selected task-specific candidate experts $\mathcal{S}_t$, we can conduct general token routing at the token level with Top-$k$ or Top-$p$ routing as described in~\autoref{topk} and~\autoref{topp}, respectively. The final routing policy can be written as $\sum^{|\mathcal{S}^t|}_{i}\sum^{|\mathcal{E}|}_{j}{g}^t_{i} \cdot {g}_{j} \cdot e_{i,j}(\mathbf{x})$.

However, current expert selection at the token level only depends on the current token representation without considering the complete word meaning and global context. Generally, the global context grasp the overall situation and might know which token is difficult or not. Thus, the context can schedules the token to route to suitable experts in a global view. To this end, we propose context-responsive experts selection at the token level.

\subsection{Context-Responsive Routing}
\label{sec:ces}
As shown in 3) of~\ref{fig.2}, before the experts selection at the token level, we explicitly inject the context into the token representation. Since it is hard to determine for a code-mixed input or a query that can belongs to several domains, which thus limits its application in practice. Therefore, we use the averaged context representation as the context representation. $\mathbf{H}_{ctx}=\frac{1}{|Y_{1:t-1}|}\sum_{t=1}^{|Y_{1:t-1}|}\mathbf{h}^{L}_{t}$ where $\mathbf{h}$ is the hidden state of the $t$-th token at the $L$ layer. At each decoding step, the context used for each input token is the prefix of the generated next token. That is, the context is dynamically updated with decoding. 

To effectively incorporate the context representation into the routing process, we design a gate to dynamically control the contribution of these information:
\begin{equation}\nonumber
  \label{gate2}
  \begin{split}
  \mathbf{x}_{i} &=  g \odot \mathbf{x}_i  + (1-g) \odot \mathbf{H}_{ctx}, \\
  {g} &= \sigma([\mathbf{x}_i; \mathbf{H}_{ctx}]\mathbf{W}^g+\mathbf{b}^g),
  \end{split}
\end{equation}
where $\mathbf{W}^g$ and $\mathbf{b}^g$ are the trainable parameters.

In this manner, we can inject the context information into each token to accurately assign specialized experts to suitable tokens and unlock the potential of the MoE.

\subsection{Training Objectives}
\label{sec:tt}
Our training objectives consist of four parts in Top-$k$ routing (translation loss $\mathcal{L}_\text{NMT}$, task prediction loss $\mathcal{L}_\text{tp}$, load balance loss at the task level $\mathcal{L}_{bd}$ and token level $\mathcal{L}_{bt}$) and five parts in Top-$p$ routing ($\mathcal{L}_\text{NMT}$, $\mathcal{L}_\text{tp}$, $\mathcal{L}_{bd}$, $\mathcal{L}_{bt}$ and dynamic routing loss $\mathcal{L}_{d}$ in Top-$p$).

\paragraph{Task Prediction Loss.}

The task prediction loss is defined as:
\begin{equation}
  \mathcal L_{tp} =  \min (- \log(\mathcal{P}^t[t_g])),
\end{equation}
where $t_g$ is the golden class label, $\mathcal{P}^t$ is calculated as in Eq.~\ref{domain_softmax}.

\paragraph{Load Balance Loss at the Task Level.}
To fully encourage the experts to specialize in certain tasks, we propose a task-guided routing loss. It aims to achieve the number of tasks processed by different experts to be roughly the same. Therefore, the task-guided routing loss is defined as:

\begin{equation}
    \mathcal{L}_{bd} = N * \sum_{i=1}^N{F_i^t * Q_i^t},
\end{equation}
where $F_i^t$ represents the proportion of tasks selecting expert $e_i$, and $Q_i^t$ denotes the proportion of the router's probability allocated to expert $e_i$. For $K$ tasks, $F_i^t$ and $Q_i^t$ are defined as: $F_i^t = \frac{1}{K}\sum_{j=1}^K{1 \{e_i \in \mathcal{S}^{t,j}\}}$ and $Q_i^t = \frac{1}{K}\sum_{j=1}^K{P_i^j}$ where $\mathcal{S}^{t,j}$ is the set of activated experts for task $j$, which is calculated by Equation~\ref{eq:S}, and $P^j$ is the probability of selecting each experts for task $j$, calculated by Equation~\ref{eq:P}.

\paragraph{Load Balance Loss at the Token Level.} 
This loss is the similar to the vanilla load balance loss with a slight difference that the experts candidates is pre-selected by the hierarchical task-guided routing. Formally, the load balance loss at the token level is defined as:
\begin{equation}
    \mathcal{L}_{bt} = |\mathcal{S}^{t}| * \sum_{k=1}^{|\mathcal{S}^{t}|}{F_k^b * Q_k^b},
\end{equation}
where $F_k^b$ represents the proportion of tokens selecting expert $e_k$, and $Q_k^b$ denotes the proportion of the router's probability allocated to expert $e_k$. For a sequence comprising $M$ tokens, $F_k^b$ and $Q_k^b$ are defined as: $F_k^b = \frac{1}{M}\sum_{\ell=1}^M{1 \{e_k \in \mathcal{S}^{t,\ell}\}}$ and $Q_k = \frac{1}{M}\sum_{\ell=1}^M{P_k^\ell}$ where $\mathcal{S}^{t,\ell}$ is the set of activated experts for token $\ell$ from pre-selected experts set $S^{d,\ell}$, which is calculated by Equation~\ref{eq:S}, and $P^\ell$ is the probability of selecting each experts for token $\ell$, calculated by Equation~\ref{eq:P}. 

\paragraph{Dynamic Routing Loss in Top-$p$.}
The dynamic routing loss aims to prevent dynamic routing from using too many parameters to cheat and losing its ability to selectively choose experts. Therefore,~\citet{huang-etal-2024-harder} introduce a constraint on $\mathbf{P}$ and minimize the entropy of the distribution $\mathbf{P}$, ensuring that every token can focus on as less specific experts as possible, which is formalized as: 
\begin{equation}
    \mathcal{L}_{topp} = - \sum_{i=1}^N {P_i * log (P_i)}.
\end{equation}

\paragraph{Final Loss.}
Our approach can be flexibly applied in Top-$k$ and Top-$p$ routing strategies. Therefore, we have two final loss functions for Top-$k$ and Top-$p$, respectively. It is a combination of the translation loss, task prediction loss, load balance loss at the task level and at the token level, and dynamic loss:
\begin{equation}
    \mathcal{J}_{topk} = \mathcal{L}_\text{NMT} + \alpha \mathcal{L}_{dp} + \beta \mathcal{L}_{bd} + \gamma \mathcal{L}_{bp},
\end{equation}
\begin{equation}
    \mathcal{J}_{topp} = \mathcal{J}_{topk} + \delta \mathcal{L}_{topp},
\end{equation}
where $\alpha$, $\beta$, $\gamma$, and $\delta$ are hyper-parameters to adjust the contribution among these loss functions, respectively. 

\section{Experiments}

\subsection{Datasets and Metric}
\noindent\textbf{Datasets.}
We use the multi-domain translation dataset proposed by~\citet{KoehnK17}. The dataset mainly covers five diverse domains: IT, Koran, Law, Medical, and Subtitles, which are available in OPUS~\cite{AulamoT19}. Following previous work~\cite{gu-etal-2022-continual,liang-etal-2024-continual}, we use the new data splitting released by~\citet{aharoni-goldberg-2020-unsupervised}, and perform German to English translation (De$\rightarrow$En). We use the OPUS-16 for multilingual translation following~\cite{zhao2024sparse}. The OPUS-16 comes from OPUS-100~\cite{zhang-etal-2020-improving}, which includes 16 languages (8 high resource (> 0.9M), 4 medium resource, and 4 low resource (< 0.1M)). Please refer to~\autoref{tab:data} of~\autoref{sec:appendix} for detailed data statistics.

\begin{table*}[]
\centering
\begin{tabular}{llcccccc}
\toprule
& \textbf{Models} & \textbf{IT} & \textbf{Koran} & \textbf{Medical} & \textbf{Law}	& \textbf{Subtitles} & \textbf{Avg.} \\ \midrule
\multirow{1}{*}{\emph{Dense}} 
&SFT-3B &40.65 & 20.4 & 51.4 & 54.8 & 28.33 & 39.12 \\\hline
\multirow{6}{*}{\emph{MoE}} 
&Trim-MoE (Top-$1$) &40.03 & 19.55 & 51.80 & 55.83 & 25.50 &38.54 \\
&Trim-MoE (Top-$2$)  & \underline{45.10} & \underline{22.68} & 51.84 & 57.12 & \underline{29.02} & 41.15\\
&Trim-MoE (Top-$p$, $p$=0.5) & 39.39 & 19.21 & \underline{55.67} & \underline{60.18} & \textbf{29.21} & 40.73\\

\cdashline{2-8}[4pt/2pt]
&THOR-MoE (Top-$1$)          &43.96$^{\ddagger}$ & 21.93$^{\ddagger}$ & 52.52$^{\dagger}$ & 56.25 & 28.41$^{\ddagger}$ & 40.61$^{\ddagger}$ \\
&THOR-MoE (Top-$2$)          &\textbf{46.00}$^{\dagger}$ & \textbf{23.35} & \textbf{55.79}$^{\ddagger}$ & \textbf{61.06}$^{\ddagger}$ & 28.23 & \textbf{42.89}$^{\ddagger}$ \\
&THOR-MoE (Top-$p$, $p$=0.5)        &44.63$^{\ddagger}$ &22.53$^{\ddagger}$ & 53.58 & 58.65 &27.99 &\underline{41.48}$^{\dagger}$ \\
\bottomrule
\end{tabular}
\caption{BLEU score on multi-domain translation benchmarks with decoder-only architecture. ``$^{\dagger}$'' and ``$^{\ddagger}$'' denote that statistically significant better than the best result of the counterpart (\emph{e.g.}, THOR-MoE (Top-2) vs. Trim-MoE (Top-2)) with t-test {\em p} \textless \ 0.05 and {\em p} \textless \ 0.01 hereinafter, respectively. The best and second best results are \textbf{bold} and \underline{underlined}, respectively.}
\label{tab:domain}
\end{table*}

\noindent\textbf{Metric.}
For a fair comparison, we follow previous work~\cite{gu-etal-2022-continual,zhao2024sparse} and adopt the 4-gram case-sensitive BLEU with the SacreBLEU tool\footnote{BLEU+case.mixed+numrefs.1+smooth.exp+tok.13a+\\version.1.4.13}~\cite{post-2018-call} and report the statistical significance test~\cite{koehn-2004-statistical}. For multilingual translation, the scores encompass the average ratings across all language pairs, such as English$\rightarrow$Any (En$\rightarrow$XX), and Any$\rightarrow$English (XX$\rightarrow$En) on the OPUS-16 dataset.

\subsection{Implementation Details}

In multi-domain translation, we use decoder-only Transformer architecture. Specifically, we use Qwen1.5-MoE-A2.7B~\cite{qwen_moe}. It has 14.3B parameters (60 non shared experts + 4 shared experts) in total and 2.7B activated parameters (4 non shared experts + 4 shared experts) during runtime. Due to limited GPU resource, we used a trimmed version, which has 3.5B parameters (8 non shared experts + 4 shared experts) in total and 2.3B activated parameters (2 non shared experts + 4 shared experts). To keep its capacity, we initialize the trimmed model with Qwen1.5-MoE-A2.7B and denote is as Trim-MoE. During training, we use Llama-Factory~\cite{zheng-etal-2024-llamafactory} to instruct-tune LLMs. All LLMs are tuned on an 8$\times$NVIDIA A100 GPUs (40G) with 1e-5 learning rate. We set gradient accumulation to 16 and batch size to 1, which gives us 2*8*16*1 batch in total. We use the DeepSpeed optimization~\cite{rasley2020deepspeed}, and set ZeRO-3 optimization. Following~\citet{qin2024o1}, we set the number of training epochs to 3. We set hyper-parameters $\alpha$, $\beta$, $\gamma$, and $\delta$ to 1e-2, 1e-2, 1e-2, and 1e-4, which are determined by a grid search.

In multilingual translation, we follow~\cite{zhao2024sparse} and compare our method with the Transformer-Base model (as Dense) and its MoE variants that have 6 encoder and decoder layers, 32 experts. The input and hidden dimensions of all feed-forward networks are 512 and 2048. We set the training processes to 35K iterations with a learning rate of 5e-4, which follows the inverse square root with 4,000 warm-up steps. To keep balanced training, we use temperature-based data sampling strategy with temperature 1.5. We set hyper-parameters $\alpha$, $\beta$, $\gamma$, and $\delta$ to 1e-2, 5e-2, 5e-2, and 1e-4, which are determined by a grid search.

\subsection{Comparison Models}
Our comparison models mainly include two types: \textbf{Decoder-only} and \textbf{Encoder-Decoder} based. 

\noindent\textbf{Decoder-only.} We fine-tune two dense models based on Qwen2.5-3B~\cite{qwen2.5}, which have similar parameters with the activated parameter and denoted as SFT-3B. Besides, based on the Trim-MoE-A2.3B model, we also fine-tuned three MoE models with Top-1, Top-2, and Top-$p$ routing strategy.

\noindent\textbf{Encoder-Decoder.} There are three types of comparison models: vanilla dense model, vanilla MoE model, and enhanced MoE models via language knowledge. We use the Transformer-Base model as Dense model. For vanilla MoE model, the Switch Transformer~\cite{fedus2022switch} with a top-1 token-based routing (as ST-MoE) and GShard~\cite{lepikhin2021gshard} with a top-2 token-based routing (as GS-MoE) are adopted. For Language-specific MoE models, the LS-MoE with fixed routing inspired by~\cite{pires-etal-2023-learning}, assigning 2 non-overlapping experts for tokens according to their source language in the encoder and target language in the decoder; Hybrid-MoE~\cite{kudugunta-etal-2021-beyond-distillation}, with a top-2 token routing in the encoder and a top-2 target language routing in the decoder side; Residual-MoE~\cite{elbayad-etal-2023-fixing,rajbhandari2022deepspeedmoeadvancingmixtureofexpertsinference,zhang2021share} that augments each MoE layer with a shared feed-forward network through a binary gate function; Lingual-MoE~\cite{zhao2024sparse} stands for an MoE model with linguistic-guided routing and dynamic expert allocation, where the first-level language router selects the top 8 experts and the second-level token router activates the dynamical number of experts. 

\begin{table*}[]
\centering
\scalebox{0.85}{
\begin{tabular}{llccccccccc}
\toprule
&\multirow{2}{*}{\textbf{Models}} &  \multicolumn{4}{c}{\textbf{En$\rightarrow$XX}}  &  \multicolumn{4}{c}{\textbf{XX$\rightarrow$En}} & \multirow{2}{*}{\textbf{Avg.}} \\
\cmidrule(lr){3-6} \cmidrule(lr){7-10}
&  & high & medium& low &  Avg1. & high & medium& low &  Avg2.  \\
\midrule
\multirow{1}{*}{\emph{Dense}} 
&Transformer-base & 25.37 &39.12 &14.78	&26.16	&28.81 &39.24 &24.21 &30.27 & 28.21\\\hline
\multirow{9}{*}{\emph{MoE}}
&GS-MoE  &25.55 &40.71 &18.97 &27.70 &28.77 &41.70 &29.25 &32.12 &29.91\\
&ST-MoE           &26.55 &43.04 &20.23 &29.09&\underline{29.94} &\underline{44.00} &30.94 &33.71 &31.40 \\\cdashline{2-11}[4pt/2pt]
&LS-MoE  &20.06 &37.31 &11.72&22.29  &23.69 &42.99 &32.01 &30.60 & 26.44\\
&Hybird-MoE  &24.56 &31.70 &13.38 &23.55 &29.59 &41.74 &27.81 &32.18 &27.85 \\
&Residual-MoE           &26.52 &43.49 &21.58 &29.53 &29.84 &43.94 &31.25 &33.72 &31.62\\
&Lingual-MoE   &\underline{27.11} &\underline{46.24} &\underline{23.34} &\underline{30.95}  &29.87 &43.49 &\underline{32.02} &\underline{33.81} &\underline{32.38}\\\cdashline{2-11}[4pt/2pt]

&THOR-MoE (Top-1)	&27.21	&46.37	&23.88	&31.17	&29.68	&44.01	&32.58	&33.99	&32.57\\
&THOR-MoE (Top-2)	&\textbf{27.88}$^{\dagger}$	&\textbf{47.29}$^{\dagger}$	&\textbf{24.85}$^{\ddagger}$	&\textbf{31.98}$^{\dagger}$	&\textbf{30.38}$^{\dagger}$	&\textbf{44.52}$^{\dagger}$	&\textbf{33.26}$^{\dagger}$	&\textbf{34.64}$^{\dagger}$	&\textbf{33.31}$^{\dagger}$ \\

&THOR-MoE (Top-$p$)         &{27.63}$^{\dagger}$	&{46.82}$^{\dagger}$	&{24.15}$^{\dagger}$	&{31.55}$^{\dagger}$  &{29.97}	&{44.18}$^{\dagger}$ &{32.96}$^{\dagger}$ & {34.26} & {32.91}$^{\dagger}$\\

\bottomrule
\end{tabular}}
\caption{Averaged BLEU scores on multilingual translation with encoder-decoder architectures. ``$^{\dagger}$'' denotes that statistically significant better than Lingual-MoE with t-test {\em p} \textless \ 0.05.}
\label{tab:language}
\end{table*}

\section{Main Results}
Table~\ref{tab:domain} shows main results on the multi-domain translation benchmark with decoder-only architecture. Table~\ref{tab:language} presents main results on the multilingual translation with encoder-decoder architecture. For a fair comparison, all models are trained and assessed on the OPUS-16 dataset with Transformer-Base as the backbone architecture. 

\subsection{Results on Multi-Domain Translation}
Table~\ref{tab:domain} shows that the Trim-MoE-based models generally surpass the dense one. For example, the Trim-MoE (Top-2) outperforms SFT-3B model by average 2 BLEU scores where the activated parameters is less than dense model (2.3B vs. 3B), showing the effectiveness of MoE model. Furthermore, the proposed THOR-MoE significantly and consistently surpasses the counterpart of Trim-MoE-based one. For instance, the THOR-MoE (Top-2) outperforms the Trim-MoE (Top-2) by averaged 1.74 BLEU scores. This clear advantage confirms the superiority of incorporating domain knowledge and context knowledge into the routing. What's important, the proposed approach is compatible with the Top-$k$~\cite{shazeer2017} and Top-$p$~\cite{huang-etal-2024-harder} routing strategies, validating the generalization of THOR-MoE, which can be a plug-and-play module. Besides, we find that the scores on the same domain (\emph{e.g.}, Law) change largely. The reason may be that the data distribution is unbalanced. We also list the results in terms of COMET~\cite{rei-etal-2020-comet} score in Table~\ref{tab:domain-comet} and we can conclude the similar findings.


\begin{table}[!t]
\centering
\small
\scalebox{0.83}{
\setlength{\tabcolsep}{0.9mm}{
\begin{tabular}{@{}lccccccc@{}}
\toprule
\textbf{Models} & \textbf{IT} & \textbf{Koran} & \textbf{Medical} & \textbf{Law}	& \textbf{Subtitles} & \textbf{Avg.} \\ \midrule
SFT-3B &86.39	&73.17	&85.43	&87.19	&78.52	&82.14 \\\hline
THOR-MoE (Top-$p$)        &87.31	&74.09	&86.09	&87.88	&79	 &82.87 \\
\bottomrule
\end{tabular}}}
\caption{COMET scores on multi-domain translation benchmarks with decoder-only architecture.}
\label{tab:domain-comet}
\end{table}

\subsection{Results on Multilingual Translation}
Table~\ref{tab:language} summarizes the results and we can conclude several observations:

\noindent \textbf{THOR-MoE vs. Dense and Vanilla MoE Baselines.} The THOR-MoE substantially surpasses the Dense and vanilla MoE multilingual translation baselines with a large margin. Specifically, compared with Dense and ST-MoE, the THOR-MoE (Top-$p$) achieves \{5.39\%, 3.99\%, 4.70\%\} and {2.46\%, 0.55\%, 1.51\%} averaged improvement in terms of BLEU score for Avg1., Avg2., and All Avg., respectively. This significant margin demonstrates the effectiveness of hierarchical language-guided routing and context-responsive routing.

\noindent \textbf{THOR-MoE vs. Language-Guided MoE models.} The THOR-MoE also consistently outperforms language-guided MoE models, including LS-MoE, hybrid-MoE, Residual-MoE, and Lingual-MoE. It shows again that the superiority of hierarchical language-guided routing and context-responsive routing. In detail, the Lingual-MoE performs the highest in baselines and it also employs a hierarchical language-group-guided and dynamical routing, which introduces the hard language id embedding and fails to consider code-mixed cases and that the token in different language can be the same (\emph{e.g.}, `Internet’ in German and English). In contrast, THOR-MoE applies a mixed language representation where the language ids are predicted, which comprehensively incorporates the languages knowledge. Furthermore, the dynamical routing in Lingual-MoE does not incorporate the context information. The context generally knows which token is difficult or not in a global view. The THOR-MoE fully considers the above issues and obtains better results.

\begin{table}[]
\centering
\resizebox{0.98\columnwidth}{!}{
\begin{tabular}{>{\color{black}} l >{\color{black}} c >{\color{black}} c }
\toprule
 & \textbf{Multi-Domain} &\textbf{Multilingual} \\
\midrule
THOR-MoE (Top-$p$)  & 41.48 & 32.91  \\ \hdashline
~\emph{w/o} hierarchical task-guided routing  & 40.16 & 31.57  \\
~\emph{w/o} context-responsive routing  & 40.74 & 32.12  \\
\bottomrule
\end{tabular}}
\caption{Ablation Study with Avg. results.}
\label{tab:ablation}
\end{table} 

\section{Analysis}

\subsection{Ablation Study}
We conduct ablation studies to investigate how well hierarchical context-responsive routing of THOR-MoE works. We conclude two findings from the results in~\autoref{tab:ablation}.

(1) ``\emph{w/o} hierarchical task-guided routing'': \emph{i.e.}, without using any task-related routing and decaying to vanilla manner (select experts from full set of experts), the model performance greatly degrades on both translation tasks. It shows the necessity of using task-guided routing in a hierarchical manner. 

(2) ``\emph{w/o} context-responsive routing'': the model performance becomes worse on both tasks when removing context. This shows that our context-responsive routing indeed can enhance the routing effectiveness and guarantee the token can be assigned to suitable and specialized experts with the indication of context, which thus benefits the model performance on both translation tasks. 

\begin{table}[]
\centering
\resizebox{0.98\columnwidth}{!}{
\begin{tabular}{>{\color{black}} l >{\color{black}} c >{\color{black}} c }
\toprule
 & \textbf{Multi-Domain} &\textbf{Multilingual} \\
\midrule
baseline (\emph{w/o} any task Rep.)  & 40.74 & 32.12  \\\hdashline
Non-Mixed Rep.  & 40.95 & 32.44  \\
Golden Rep.  & 41.32 & 32.78  \\
THOR-MoE (Mixed manner)  & 41.48 & 32.91  \\ 
\bottomrule
\end{tabular}}
\caption{The Avg. results of different representation (Rep.) manner during hierarchical task-guided routing.}
\label{tab:ablation2}
\end{table}

\subsection{Comparison among Different Task Representations}
\label{ITP}
In this section, we aim to investigate the impact of task representation in different manners. Table~\ref{tab:ablation2} shows the results where the `Non-Mixed Representation' denotes that directly uses the automatically predict task label to extract corresponding task representation and `Golden Representation' indicates using the golden task label to extract corresponding representation. We conclude that the task knowledge indeed has a positive impact on translation performance (vs. baseline). We also observe that using the automatically predicted task labels (actually the mixed task representation) shows better results than using ground truth and Non-Mixed manner in terms of Avg. BLEU scores. The reason may be that the mixed task representation has certain fault tolerance. We also analyze the accuracy of task prediction to further show its quality in Appendix~\ref{TP}.

\begin{table}[]
\centering
\resizebox{0.98\columnwidth}{!}{
\begin{tabular}{>{\color{black}} l >{\color{black}} c >{\color{black}} c }
\toprule
 & \textbf{Multi-Domain} &\textbf{Multilingual} \\
\midrule
baseline (\emph{w/o} any task Rep.)  & 40.74 & 32.12  \\\hdashline
Infuse task Rep. into token Rep.   & 40.95 & 32.44  \\
THOR-MoE (Hierarchical manner)  & 41.48 & 32.91  \\ 
\bottomrule
\end{tabular}}
\caption{The investigation of why hierarchical manner.}
\label{tab:ablation3}
\end{table} 

\subsection{Why Hierarchical Design?}
In this section, we aim to investigate the manner of using mixed task representation. Table~\ref{tab:ablation3} shows that introducing task knowledge indeed helps tokens route well and achieve improvement (vs. baseline). The hierarchical manner greatly outperforms the direct infusion with token representation before routing, proving the effectiveness of hierarchical design, which unlock the potential of MoE with the help of the mixed task knowledge. 
\textbf{\begin{figure}[t]
    \centering
    \includegraphics[width=0.49\textwidth]{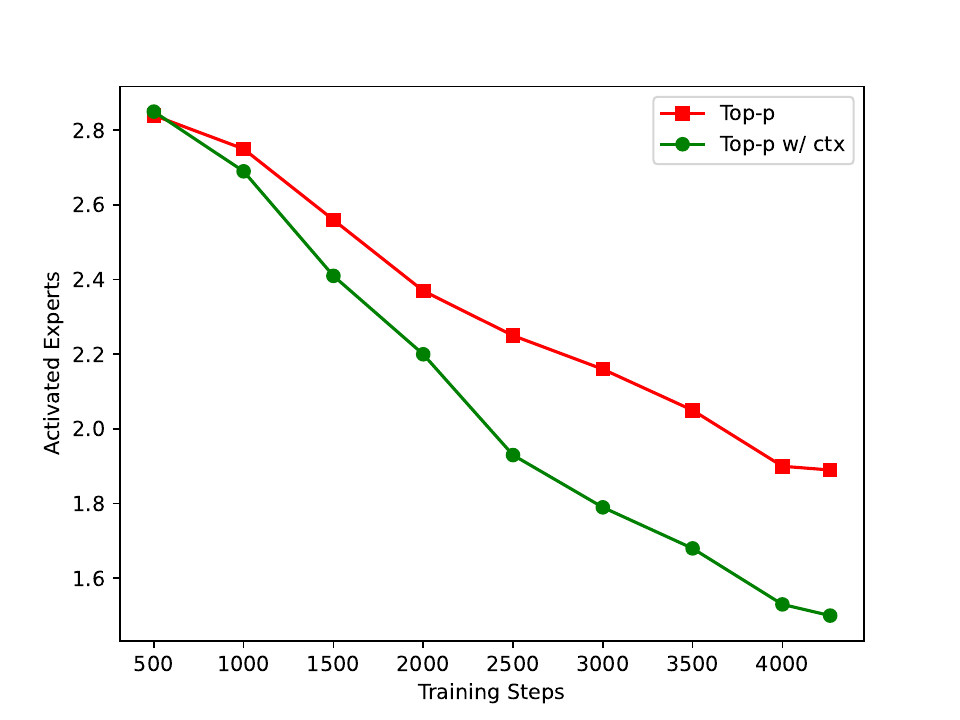}
    \caption{Average activated experts number across training steps on the multi-domain translation task.}
    \label{fig.prompts_vanilla}
\end{figure}}
\subsection{Analysis of Efficient Training and Inference}
To further explore whether our proposed method is efficient in training and inference, we calculate the average number of experts activated by the model on multi-domain translation tasks. Figure~\ref{fig.prompts_vanilla} and Table~\ref{tab:act_experts} shows the average number of experts activated per token across various translation tasks during training and inference. The result is averaged across all the layers of transformers.

During training, we can see that our method converges faster to use less activated experts than the vanilla Top-$p$~\cite{huang-etal-2024-harder}. During inference, we can observe that across all five tasks, the number of activated experts is less than original Top-$p$ routing, averaging 1.45 activated experts (less than 22\% activated parameters) with better performance. Both findings show that the context plays a key role in guiding token routing.

\section{Related Work}
\noindent\textbf{Neural Machine Translation.} The NMT have received remarkable attention in the era of LLMs~\cite{DBLP:conf/acl/JawaharMLKALABG23}. Previous work mainly focuses on the continual learning of new domains~\cite{gu-feng-2020-investigating,gu-etal-2022-continual,liang-etal-2024-continual}, introducing ready-made task-related (linguistics) knowledge via adapter~\cite{zhang-etal-2020-improving,zhang2021share}, parameter sharing~\cite{aharoni-etal-2019-massively}, or task-specific modules~\cite{pires-etal-2023-learning}, and multilingual representation learning~\cite{yang-etal-2021-improving-multilingual}. 

\begin{table}[!t]
\centering
\small
\setlength{\tabcolsep}{0.9mm}{
\begin{tabular}{@{}lccccccc@{}}
\toprule
Domains         & \textbf{IT} & \textbf{Koran} & \textbf{Medical} & \textbf{Law}	& \textbf{Subtitles} & \textbf{Avg.} \\ \midrule
Top-$p$                 & 1.87 & 1.95 & 1.82 & 1.92 & 1.77 & 1.87             \\
Top-$p$ \emph{w/} ctx   & 1.37 & 1.61 & 1.42 & 1.56 & 1.31 & 1.45  \\
\bottomrule
\end{tabular}}
\caption{Average activated experts (AE) in different domain translation tasks. `Top-$p$ \emph{w/} ctx' denotes the proposed method to infuse the context during routing. }
\label{tab:act_experts}
\end{table}
\noindent\textbf{Mixture-of-Experts.}~\citet{DBLP:journals/neco/JacobsJNH91} first proposes the concept of MoE, which consists of a series of network sub-modules. The Sparsely-gated MoE, as a variant, which only activates a few expert networks for each input, has been shown its superiority in diverse NLP and computer vision applications\cite{DBLP:conf/iclr/ShazeerMMDLHD17,zoph2022st}. In the context of NMT, most of previous work focuses on addressing the over/under-fitting~\cite{elbayad-etal-2023-fixing} via regularization strategies~\cite{elbayad-etal-2023-fixing} or modularizing MoE~\cite{li-etal-2023-mmnmt,zhang-etal-2024-lightweight}, and incorporating task-related knowledge (language, domain \emph{etc})~\cite{kudugunta-etal-2021-beyond-distillation,zhao2024sparse,pham-etal-2023-task,gururangan-etal-2022-demix,li2023branchtrainmerge}, achieving impressive performance. Besides, there are some studies aim to improve the training/inference efficiency of routing via adaptive computation~\cite{jawahar-etal-2023-automoe} or reducing the number of activated experts~\cite{li-etal-2023-adaptive-gating,huang-etal-2024-harder}. 

Different from these prior works that directly use the task-related knowledge, we propose a hierarchical context-responsive routing method where we automatically extract corresponding domain/language knowledge and design a hierarchical network to guide the task-level routing. Besides, we aim to help each token accurately select specialized and suitable experts and thus we incorporate the context to guide the token routing rather than relying on the localized token only in existing work. In this manner, the context-responsive routing can improve the training or inference efficiency, which is compatible with previous Top-$k$ and Top-$p$ routing policies. 

\section{Conclusion}
In this paper, we propose a new hierarchical task-gudied and context-responsive routing framework for NMT. To automatically obtain the task knowledge, we propose to predict it and then use mixed task representation. Consequently, we design a hierarchical routing at the task level and the token level. Further, we propose to inject context to enhance the effectiveness of token routing. Extensive experiments on multi-domain and multilingual translation benchmarks show the superiority and generalization of our proposed approach. 

\section*{Limitations}

While we introduce the mixed domain/language and context knowledge into routing in hierarchical manner and achieve good results, there are some limitations worth considering to study in future work: (1) In this study, the design relies on the prior (the number of tasks and language groups), which may limit its extention to broader topics~\cite{li2025your}; (2) This work only conduct experiments on translation tasks and does not conduct experiments other generation or discriminative tasks.


\bibliography{custom}

\clearpage
\newpage
\appendix

\label{sec:appendix}

\begin{table}[]
\centering
\resizebox{0.98\columnwidth}{!}{
\begin{tabular}{lcccc}
\toprule
    & Tasks/Groups& Train & Valid & Test \\ \midrule
\multicolumn{1}{l}{\multirow{5}{*}{\begin{tabular}[c]{@{}l@{}}Multi-Domain\\ Translation\\ Dataset \\ (De$\rightarrow$En)\end{tabular}}} 
& IT & 0.22M & \multirow{5}{*}{2000} & \multirow{5}{*}{2000} \\
\multicolumn{1}{l}{} & Koran & 18K &  &  \\
\multicolumn{1}{l}{} & Law & 0.47M &  &  \\
\multicolumn{1}{l}{} & Medical & 0.25M &  &  \\
\multicolumn{1}{l}{} & Subtitles & 0.5M &  &  \\\midrule
\multicolumn{1}{l}{\multirow{3}{*}{\begin{tabular}[c]{@{}l@{}}Multilingual\\ Translation\\ Dataset \end{tabular}}} 
& \multirow{3}{*}{16} & \multirow{3}{*}{17,559,950} & \multirow{3}{*}{30*1000} & \multirow{3}{*}{30*1000} \\
& \\
& \\
\bottomrule
\end{tabular}}
\caption{{The data statistic of the multi-domain translation dataset and multilingual translation (OPUS-16) dataset. The number in Train/Valid/Test columns denotes the number of sentence pairs in each domain/language pair.}}
\label{tab:data}
\end{table}

\section{Task Prediction}
\label{TP}
We also evaluate the performance of the \emph{Task Predictor} to show whether the classifier can accurately predict suitable labels. The results are 82.45\% and 64.89\% for domain and language prediction, respectively. It suggests that our classifier can predict suitable labels and further provide effective mixed task representation for routing. 

\section{Comparison to Larger Model}

We implemented our proposed method based on a large-scale Qwen1.5-MoE-A2.7B (totally 14B, activated 2.7B, denoted as THOR-MoE). Besides, we also fine-tuned the vanilla Qwen1.5-MoE-A2.7B model as the baseline. The averaged COMET and BLEU results shown in Table~\ref{tab:domain-larger} demonstrate that the THOR-MoE model at the 14B level also achieve better results than fine-tuned Qwen1.5-MoE-A2.7B model in terms of BLEU and COMET scores. 

\begin{table}[!t]
\centering
\small
\scalebox{0.7}{
\setlength{\tabcolsep}{0.9mm}{
\begin{tabular}{@{}lccccccc@{}}
\toprule
\textbf{Models} & \textbf{IT} & \textbf{Koran} & \textbf{Medical} & \textbf{Law}	& \textbf{Subtitles} & \textbf{Avg.} \\ \midrule
Qwen1.5-MoE-A2.7B (SFT, Top-8)	&45.59	&23.31	&55.67	&60.18	&29.21	&42.79\\\cdashline{1-7}[4pt/2pt]
THOR-MoE (Top-p, p=0.5)	&46.76	&24.03	&55.9	&61.13	&30.29	&43.62\\\hline\hline

Qwen1.5-MoE-A2.7B (SFT, Top-8)	 &87.3	&74.24	&85.85	&87.78	&79.36	&82.91 \\\cdashline{1-7}[4pt/2pt]
THOR-MoE (Top-p, p=0.5)	&87.45	&74.58	&86.45	&88.75	&80.23	&83.49 \\
\bottomrule
\end{tabular}}}
\caption{BLEU (top block) / COMET (below block) scores on multi-domain translation benchmarks with decoder-only architecture. Note top-8 means it activates 4 shared experts and 4 non-shared experts out of 60 experts.}
\label{tab:domain-larger}
\end{table}

\section{Analysis of Domain Similarity}

Following previous work~\cite{wu-etal-2024-f}, we conduct some analysis for domain similarity. Specifically, we sum the predicted soft label and then normalize it. The vertical axis and horizontal axis are unsupervised cluster label (clustering with BERT-base and k=5) and predicted soft labels, respectively. The similarity matrix are shown in Table~\ref{tab:domain-ds}.

\begin{table}[!t]
\centering
\small
\setlength{\tabcolsep}{0.9mm}{
\begin{tabular}{@{}lccccccc@{}}
\toprule
 & \textbf{IT} & \textbf{Koran} & \textbf{Medical} & \textbf{Law}	& \textbf{Subtitles} \\ \midrule
IT	&0.833	&0.027	&0.055	&0.0219	&0.065\\
Koran	&0.012	&0.817	&0.003	&0.019	&0.149\\
Medical	 &0.111	&0.004	&0.737	&0.135	&0.013 \\
Law &0.142	&0.008	&0.032	&0.798	&0.02 \\
Subtitles	&0.058	&0.076	&0.009	&0.006	&0.851 \\
\bottomrule
\end{tabular}}
\caption{BLEU (top block) / COMET (below block) scores on multi-domain translation benchmarks with decoder-only architecture. Note top-8 means it activates 4 shared experts and 4 non-shared experts out of 60 experts.}
\label{tab:domain-ds}
\end{table}

The results show that there exist some sentences that are assigned to a cluster of another domain (i.e., hard cases). This makes sense as the mixed representation can capture such domain information, making the proposed approach works (the reason).


\end{document}